\documentclass{article}

\usepackage[preprint]{neurips_2026}

\usepackage[utf8]{inputenc} 
\usepackage[T1]{fontenc}    
\usepackage{hyperref}       
\usepackage{url}            
\usepackage{booktabs}       
\usepackage{amsfonts}       
\usepackage{amsmath}        
\usepackage{nicefrac}       
\usepackage{microtype}      
\usepackage{xcolor}         
\usepackage{graphicx}       
\usepackage{multirow}       
\usepackage{makecell}       
\usepackage{bm}             

\newcommand{\estci}[3]{#1\textsuperscript{\scriptsize\,(#2--#3)}}
\newcommand{\bestci}[3]{\textbf{#1}\textsuperscript{\scriptsize\,(#2--#3)}}
\newcommand{\real}{\mathbb{R}}

\title{An Elastic Shape Variational Autoencoder for Skeleton Pose Trajectories}

\author{
Arafat Rahman$^{1}$\thanks{Equal contribution.},
 Shashwat Kumar$^{2}$\footnotemark[1],
 Laura E. Barnes$^{1}$,
 Anuj Srivastava$^{3}$
\\
$^{1}$Systems and Information Engineering, University of Virginia, Charlottesville, VA, USA \\
$^{2}$Biomedical Engineering, Johns Hopkins University, Baltimore, MD, USA \\
$^{3}$Dept. of Applied Mathematics and Statistics, Johns Hopkins University, Baltimore, MD, USA \\
\\
\texttt{\{jgh6ds, lb3dp\}@virginia.edu}\\
\texttt{skuma118@jh.edu,} \texttt{anuj.srivastava@jhu.edu}
}

\begin{document}

\maketitle

\begin{abstract}
Deep generative models provide flexible frameworks for modeling complex, structured data such as images, videos, 3D objects, and texts. However, when applied to sequences of human skeletons, standard variational autoencoders (VAEs) often allocate substantial capacity to nuisance factors—such as camera orientation, subject scale, viewpoint, and execution speed—rather than the intrinsic geometry of shapes and their motion. We propose the Elastic Shape - Variational Autoencoder (ES-VAE), a geometry-aware generative model for skeletal trajectories that leverages the transported square-root velocity field (TSRVF) representation on Kendall’s shape manifold. This representation inherently removes rigid translations, rotations, and global scaling of shapes, and temporal rate variability of sequences, isolating the underlying shape dynamics. The ES-VAE encoder maps skeletal sequences to a low-dimensional latent space incorporating the Riemannian logarithm map, while the decoder reconstructs sequences using the corresponding exponential map. We demonstrate the effectiveness of ES-VAE on two datasets. First, we analyze skeletal gait cycles to predict clinical mobility scores and classify subjects into healthy and post-stroke groups. Second, we evaluate action recognition on the NTU RGB+D dataset. Across both settings, ES-VAE consistently outperforms standard VAEs and a range of sequence modeling baselines, including temporal convolutional networks, transformers, and graph convolutional networks. More broadly, ES-VAE provides a principled framework for learning generative models of longitudinal data on pose shape manifolds, offering improved latent representation and downstream performance compared to existing deep learning approaches.
\end{abstract}


\section{Introduction}
Skeleton trajectories have become ubiquitous in activity recognition~\citep{sun2022human}, gait recognition~\citep{fan2024skeletongait}, automatic motion assessment~\citep{hakim2019mal}, and disease severity prediction~\citep{adeli2024benchmarking}. Despite the advances in data collection, extracting meaningful features from such high-dimensional data is far from straightforward.


\begin{figure*}[!t]
    \centering
    \includegraphics[width=\linewidth]{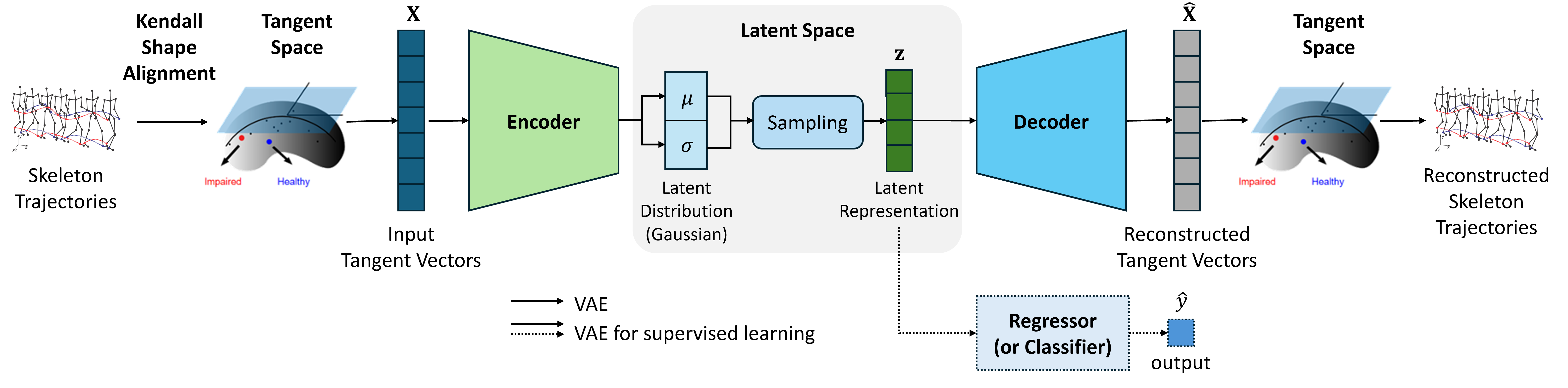}
    \caption{Overview of the Elastic Shape VAE architecture. Raw skeleton trajectories are embedded in Kendall shape space by removing translation, scale, and rotation. Temporal alignment via the transported SRVF removes rate variability. The Riemannian VAE encoder maps tangent vectors to a low-dimensional latent code $z$, and the decoder maps $z$ back through the tangent space and onto the shape manifold via the exponential map.}
    \label{fig:overview}
\end{figure*}

One approach to learn interpretable representations from high dimensional data is to use variational autoencoders (VAE)~\citep{kingma2013auto}. These have proven to be useful in applications ranging from computer vision~\citep{harvey2022conditional} to computational biology~\citep{lopez2018deep}. Applied to pose data, though, a standard VAE devotes a large part of its capacity to irrelevant dimensions, {\it e.g.}, camera orientation, sensor distance, body size and rate at which the activity is performed. These nuisance variables are unavoidable in naturalistic gait or activity recording setups where participants walk or perform activity at their own pace and starting position~\citep{van2023full}. However, they play no role in deriving inferences about activity classification or labeling. In fact, it is common in traditional shape analysis~\citep{dryden2016statistical} to remove certain nuisance transformations before applying statistical analyses. The question is how to perform this removal while involving powerful ML and AI techniques that are often geared towards Euclidean variables.

In order to develop an AI-based technique for analyzing shape sequences, while focusing directly on shapes and shape dynamics, we represent skeleton trajectories as stochastic processes in \emph{Kendall shape space}~\citep{kendall2009shape}. The representation automatically quotients out translations, rotations, and scaling of individual shapes. We pair this with the \emph{transported square-root velocity field} (TSRVF) framework~\citep{amor2015action} on shape manifold to remove the temporal rate variability (activity speed) of sequences. The resulting model, \emph{Elastic Shape VAE} (ES-VAE), is invariant to these confounders by design while focusing on shape dynamics in form of meaningful latent representations. Figure~\ref{fig:overview} shows the overview of the full architecture.

\subsection{Contributions}
\begin{enumerate}
    \item We introduce Elastic Shape VAE, a Riemannian VAE that full embeds skeleton trajectories -- represented by transported SRVFs on Kendall shape space -- into low-dimensional latent space. This representation results in complete invariance to shape translations, rotations, scaling, and speed profiles of activities.
    \item Using the unit sphere $\mathbb{S}^2$ as an example, we show that Elastic Shape VAE approximates non-geodesic, one-dimensional submanifolds better than such popular ideas Euclidean PCA, Euclidean VAE, and tangent-space PCA. This demonstrates that modeling directly on the manifold preserves nonlinear geometric structure that is lost under Euclidean approximations.
    \item On a clinical dataset of 155 participants (111 healthy, 44 stroke), we demonstrate that the learned latent dimensions: (1) map to interpretable gait features (stride length, limb stiffness, arm variability), (2) correlate highly with clinical scores, and (3) outperform other classical and deep learning approaches for predicting stroke severity.
    \item On a representative subset of activity recognition dataset, the learned latent dimensions similarly outperformed other classical and deep learning approaches in classifying different activites. 
\end{enumerate}

The remainder of this paper is organized as follows. Section~\ref{sec:related} reviews related work. Section~\ref{sec:methodology} presents the mathematical framework. Section~\ref{sec:experiments} describes the experimental setup. Section~\ref{sec:results} presents the results. Section~\ref{sec:discussion} discusses the findings and limitations, and Section~\ref{sec:conclusion} concludes.

\section{Related Literature}
\label{sec:related}

\noindent {\bf Riemannian and Geometric Deep Learning}:
Deep generative models have been pushed beyond Euclidean space in several ways. \citet{miolane2020learning} introduced the Riemannian VAE, where the decoder maps latent codes through the tangent space at a base point and then onto the manifold via the exponential map. The network therefore only has to learn the residual, not the full manifold geometry. \citet{chadebec2022geometric} used inverse posterior covariances to define a Riemannian metric in the VAE latent space and then used geometry-aware sampling to improve generation compared to standard Gaussian sampling. On the shape-modeling side, \citet{bone2019learning} built diffeomorphic autoencoders for neuroimaging atlas construction. \citet{dummer2024rda, dummer2023rsa} developed Riemannian diffeomorphic autoencoders with implicit neural representations. \citet{gatti2024shapemed} benchmarked neural shape models on 3D femur data, where hybrid explicit-implicit models beat classical statistical shape models. \citet{fu2025manifoldformer} combined Riemannian VAEs with geodesic-aware Transformers for EEG decoding. All of these operate on general manifolds. We instead work on \emph{Kendall shape space}, which is purpose-built for skeleton data: translation, rotation, and scale invariance come from the geometric construction itself, so there is no need for data augmentation or learned invariances. \citet{vadgama2022kendall} used a Kendall pre-shape latent representation, learning landmark-like codes from images using VAE.
In contrast, our inputs are skeletons, whose joints are already landmarks, so we model them directly in Kendall shape space.

\noindent {\bf Shape Analysis of Skeletal Data}:
Several works have been done on the shape analysis of skeleton trajectories. \citet{amor2015action} used the transported SRVF framework on Kendall shape space for rate-invariant action recognition. \citet{hosni20183d} extended the idea to 3D gait recognition with functional PCA on Kendall shape space. \citet{friji2021geometric} paired Kendall shape representations with a CNN-LSTM classifier for skeleton action recognition and added a differentiable transformation layer to handle both rigid and non-rigid normalizations. These are all discriminative approaches (classification). We instead integrate shape-space invariances into a VAE, yielding an interpretable latent embedding of skeleton shape trajectories that suppresses nuisance variation. While there have been several papers on AI-based encoding of {\it static} shapes, this is the first effort to encode full shape trajectories in a manner that is invariant to all spatio-temporal nuisance variables. Additional related works are given in Appendix Section~\ref{sec:related_works_appendix}.


\section{Methodology}
\label{sec:methodology}
In this section, we utilize the geometry of Kendall's shape representation and develop low-dimensional VAE-based representations of human skeleton and their evolutions. 

\subsection{Kendall Shape Representations}
\label{sec:kendall}

Each human skeleton is made of $k$ landmarks in $\real^m$ where $m=3$ is the ambient dimension. 
We represent a skeleton in a matrix form $X \in \mathbb{R}^{k \times m}$. The inner product between any two configurations $X_1, X_2 \in \mathbb{R}^{k \times m}$ is $\langle X_1, X_2 \rangle = \mathrm{tr}(X_1^\top X_2)$. Following \citet{kendall2009shape}, we define the \emph{shape} of a skeleton as the geometric information that remains after removing position, orientation, and size. The full process is described below. 

\textbf{Removing translation}: To remove translation, we simply center each configuration by subtracting the column means, yielding the centered configuration space:
\begin{equation}
V_m^k = \left\{ X \in \mathbb{R}^{k \times m} : \sum_{i=1}^{k} X_{i,:} = \mathbf{0} \right\}.
\end{equation}

\textbf{Removing scale.} We normalize each centered configuration to unit Frobenius norm, yielding the \emph{preshape space}:
$S_m^k = \left\{ X \in V_m^k : \| X \|_F = 1 \right\}$.
The unit-norm constraint endows the preshape space with a spherical geometry. We impose the standard Euclidean metric as a Riemannian metric on $S_m^k$ which leads to the geodesic distance $d_S(X_1, X_2) = \cos^{-1}(\langle X_1, X_2 \rangle)$.

\textbf{Removing rotation.} Even after removing position and scale, orientation remains as a nuisance variable. The special orthogonal group $\mathrm{SO}(m) = \{ R \in \mathbb{R}^{m \times m} : R^\top R = I,\, \det(R) = 1 \}$ acts on preshape space by right multiplication: $X \mapsto XR$. This action preserves the preshape metric, since for tangent vectors $u, v \in T_X S_m^k$:
$\langle uR, vR \rangle = \mathrm{tr}(R^\top u^\top v R) = \mathrm{tr}(u^\top v) = \langle u, v \rangle$.

We define an equivalence relation: $X_1 \sim X_2$ if and only if $X_2 = X_1 R$ for some $R \in \mathrm{SO}(m)$. \emph{Kendall shape space} is the quotient:
$\Sigma_m^k = S_m^k / \mathrm{SO}(m) = \{[X] | X \in S_m^k\}$, where an equivalence class is defined as $[X] = \{ XO: O \in SO(m)\}$.

Away from singularities, the quotient map $\pi: S_m^k \to \Sigma_m^k$, given by $\pi(X) = [X]$,  is a Riemannian submersion and induces a metric on $\Sigma_m^k$ from $S_m^k$. The geodesic distance between two shapes $[X_1], [X_2] \in \Sigma_m^k$ is:
\begin{equation}
d_{\Sigma_m^k}([X_1], [X_2]) = \inf_{R \in \mathrm{SO}(m)} d_{S}(X_1, X_2 R),
\label{eq:shape_dist}
\end{equation}
where $d_S$ is as defined earlier. The optimal rotation $R^*$ is found via Orthogonal Procrustes Analysis (OPA).

\textbf{Exponential and logarithmic maps.}
Given a base point $\nu \in S_m^k$ and a tangent vector $w \in T_\nu S_m^k$, the exponential map on the preshape sphere is:
\begin{equation}
\mathrm{Exp}_\nu(w) = \cos(\|w\|_F)\,\nu + \sin(\|w\|_F)\,\frac{w}{\|w\|_F}.
\end{equation}
The logarithmic map, which maps a point $X \in S_m^k$ to a tangent vector at $\nu$, is given by:
\begin{equation}
\mathrm{Log}_\nu(X) = \frac{\theta}{\sin \theta}\left(X - \cos(\theta)\,\nu\right),\ \mbox{where}\ \theta = \cos^{-1}(\langle \nu, X \rangle).
\end{equation}

\subsection{Transported SRVF and Temporal Alignment}
\label{sec:tsrvf}

A skeleton trajectory is a curve $\beta: [0,1] \to \Sigma_m^k$ parameterized by time. The set of all such smooth trajectories is given by $\mathcal{M}$. For any point $\mu \in {\cal M}$, the tangent space $T_{\mu}{\cal M} = \prod_{t} T_{\mu(t)}\Sigma_m^k$. The log and exponential maps for full trajectories are defined using the their analogs at each time $t$. That is: $\mathrm{Log}_{\mu}(\beta_n) = \{\mathrm{Log}_{\mu(t)}(\beta_n(t))~:~t \in [0,1]\}$ and 
$\mathrm{Exp}_\mu(v) = \{\mathrm{Exp}_{\mu(t)}(v(t))~:~ t \in [0,1]\}$.

Different participants may walk at different speeds, introducing rate variability. We handle this using the \emph{transported square-root velocity function} (TSRVF)~\citep{amor2015action}.

For a trajectory $\beta \in \mathcal{M}$, we compute the covariant derivative $\dot{\beta}(t)$ via:
$\dot{\beta}(t) \approx \frac{\mathrm{Log}_{\beta(t)}(\beta(t+\Delta t))}{\Delta t}$,
and parallel transport the resulting tangent vectors to a common reference point $c \in \Sigma_m^k$. As common reference point we used Fr\'{e}chet mean. The TSRVF $q:[0,1] \to T_c\Sigma_m^k$ is then:
\begin{equation}
q(t) = \frac{(\dot{\beta}(t))_{\beta(t) \to c}}{\sqrt{\|(\dot{\beta}(t))_{\beta(t) \to c}\|_2}},
\end{equation}
where the subscript denotes parallel transport from $\beta(t)$ to $c$.

\textbf{Temporal alignment.} Let $\Gamma$ denote the set of orientation-preserving diffeomorphisms of $[0,1]$ (reparameterizations). For a trajectory $\beta$ and a $\gamma \in \Gamma$, the composition $\beta \circ \gamma$ denotes the same set of points, but traversed at a rate governed by $\gamma$. For any two trajectories $\beta_n$ and $\mu$, the optimal time warping is:
\begin{equation}
\gamma_{n}^* = \arg\min_{\gamma \in \Gamma} \int_0^1 \left\| q_{\mu}(t) - q_{\beta}(\gamma(t))\sqrt{\dot{\gamma}(t)} \right\|^2_2 dt,
\end{equation}
which is solved via efficient Dynamic Programming Algorithm (DPA)~\citep{bertsekas2012dynamic}. The aligned trajectory is $\tilde{\beta}_i = \beta_i \circ \gamma_i^*$.

\subsection{Mean Trajectory and Registration}
\label{sec:alignment}

Given a collection of skeleton trajectories $\{\beta_n(t)\}_{n=1}^N$ on Kendall shape space, we iteratively compute their Fr\'{e}chet mean $\mu \in {\mathcal M}$ and simultaneously align and register all trajectories to this mean. 
This process iterates, alternating between (i) rotationally aligning each shape $\beta_n(t)$, for each $n$, to the current mean $\mu(t)$ via OPA, (ii) temporally aligning each trajectory $\beta_n$ to the current mean $\mu$ via TSRVF optimization, and (iii) updating the mean, until convergence. While steps (i) and (ii) are discussed earlier, we explain the updating of mean.  For each $n$, we first compute the shooting vectors: 
$v_n = \mathrm{Log}_{\mu}(\beta_n)$ and their average  $\bar{v} = \frac{1}{N}\sum_{n=1}^N v_n$. We use this mean to update the current mean according to:
$\mu \leftarrow \mathrm{Exp}_{\mu}(\epsilon \, \bar{v})$,
where $\epsilon$ is a step size. 

The output of this process is the mean trajectory $\mu$ and a set of aligned and registered trajectories $\{\tilde{\beta}_n\}$ with corresponding tangent vectors  $v_n \in T_\mu \mathcal{M}$ (equivalently, $\{v_n(t) \in T_{\mu(t)}\Sigma_m^k : t \in [0,1]\})$.

\subsection{Riemannian VAE on Shape Space}
\label{sec:rvae}

We use a latent space encoding to impose probability distribution on the trajectory space ${\cal M}$. 
For a latent space $\mathbb{R}^L$, let $f_\theta: \mathbb{R}^L \to T_\mu \mathcal{M}$ denote a neural network mapping from the latent space to the tangent space at $\mu$. Then, each aligned trajectory $\tilde{\beta}_n$ is modeled as a random sample according to:
\begin{equation}
\tilde{\beta}_n \sim \mathcal{N}_{\mathcal{M}}\!\left(\mathrm{Exp}_\mu\!\big(f_\theta(z_n)\big),\; \sigma^2\right),
\end{equation}
where $z_n \sim \mathcal{N}(\mathbf{0}, \mathbf{I}_L)$ denote a latent code of dimension $L$. 

\textbf{Encoder.} The encoder takes as input the tangent vector representation $v_n = \mathrm{Log}_\mu(\tilde{\beta}_n)$ and produces the parameters of an approximate posterior:
\begin{equation}
h_\phi(z_n \mid v_n) = \mathcal{N}\!\left(\mu_\phi(v_n),\, \mathrm{diag}(\sigma^2_\phi(v_n))\right),
\end{equation}
where $\mu_\phi$ and $\sigma_\phi$ are parameterized by neural networks.

\textbf{Decoder.} The decoder maps the latent code $z_n$ to a tangent vector $f_\theta(z_n) \in T_\mu \mathcal{M}$, which is then projected onto the manifold: $\hat{\beta}_n = \mathrm{Exp}_\mu(f_\theta(z_n))$.

\textbf{Riemannian Evidence Lower Bound (ELBO).} As training objective we used the Riemannian ELBO:
\begin{equation}
\begin{aligned}
\mathcal{L}(\theta,\phi) &= \mathbb{E}_{h_\phi(z \mid v_n)}\!\left[\log p_\theta(\tilde{\beta}_n \mid z)\right]-\mathrm{KL}\!\left(h_\phi(z \mid v_n) \,\|\, p(z)\right),
\end{aligned}
\label{eq:elbo}
\end{equation}
where the reconstruction term uses the squared geodesic distance on the shape manifold:
\begin{equation}
-\log p_\theta(\tilde{\beta}_n \mid z) \propto \sum_{t=1}^{T} d_{\Sigma_m^k}\!\left(\tilde{\beta}_n(t),\, \hat{\beta}_n(t)\right)^2,
\label{eq:recon}
\end{equation}
Here, $\hat{\beta}_n(t)$ is the reconstructed skeleton trajectory and the KL divergence for the diagonal Gaussian encoder against the standard normal prior is:
\begin{equation}
\mathrm{KL}_n = \frac{1}{2}\sum_{j=1}^{L}\left(\sigma_j^2(v_n) + \mu_j^2(v_n) - 1 - \log \sigma_j^2(v_n)\right).
\label{eq:kl}
\end{equation}

The total loss over a mini-batch of $N$ samples is:
\begin{equation}
\mathcal{L} = \frac{1}{N}\sum_{n=1}^{N}\left(\sum_{t=1}^{T} d_{\Sigma_m^k}(\tilde{\beta}_n(t), \hat{\beta}_n(t))^2 + \beta_{\text{kl}} \cdot \mathrm{KL}_n\right),
\label{eq:total_loss}
\end{equation}
where $\beta_{\text{kl}}$ controls the trade-off between reconstruction fidelity and latent regularization. The full derivation is given in Appendix Section~\ref{app:elbo}.

\section{Experimental Setup}
\label{sec:experiments}

\noindent {\bf Datasets}:
\label{sec:datasets}
We evaluate ES-VAE on two skeleton datasets: a clinical gait stroke dataset for fine-grained continuous prediction and laterality classification, and full NTU-60 action recognition dataset. On the stroke dataset, we predicted Tinetti Performance Oriented Mobility Assessment (POMA) scores and performed three-class classification (left-sided stroke, right-sided stroke, and healthy subject). On NTU-60, we evaluated action recognition on 60 different activities. The full dataset details are given in Appendix Section~\ref{sec:appendix_dataset}. 

\noindent {\bf Implementation Details}:
The Fr\'{e}chet mean was computed over 40 iterations with step size $\epsilon = 0.1$ and convergence tolerance $10^{-5}$. Rotational alignment was performed via OPA at each iteration, and temporal alignment via dynamic programming on the TSRVF representation. The same alignment pipeline is applied to both datasets; only the input dimensionality and the joint count differ. Details about network architecture, training, inference, and evaluation protocol are provided in Appendix Section~\ref{sec:addtional_implementation}.

\noindent {\bf Baseline Methods}:
We compare ES-VAE against a comprehensive set of baselines spanning different categories of input representations:

1. \textbf{Raw skeleton methods:} PCA + $k$-NN, VAE + $k$-NN, Temporal Convolutional Network (TCN)~\citep{lea2017temporal}, Long Short-Term Memory (LSTM)~\citep{hochreiter1997long}, Transformer~\citep{vaswani2017attention}, and Spatio-Temporal Graph Convolutional Network (ST-GCN)~\citep{yan2018spatial}, and  Hyper-Graph Convolutional Network (Hyper-GCN)~\citep{zhou2025adaptive} applied directly to unregistered skeleton sequences.

2. \textbf{Joint angle methods:} PCA on gait-normalized joint angles~\citep{cho2024stroke} + $k$-NN, and VAE on joint angles + $k$-NN for stroke dataset as joint angles were available in that dataset only and not in NTU-60 dataset.

3. \textbf{Tangent vector methods:} Tangent PCA + $k$-NN, which applies PCA in the tangent space at the Fr\'{e}chet mean after Kendall shape alignment (the same alignment used by ES-VAE, but with a linear dimensionality reduction instead of VAE).

\section{Experimental Results}
\label{sec:results}

\subsection{Results on Synthetic Data}
\label{sec:synthetic}

\begin{figure}[!t]
    \centering
    \includegraphics[width=0.60\linewidth]{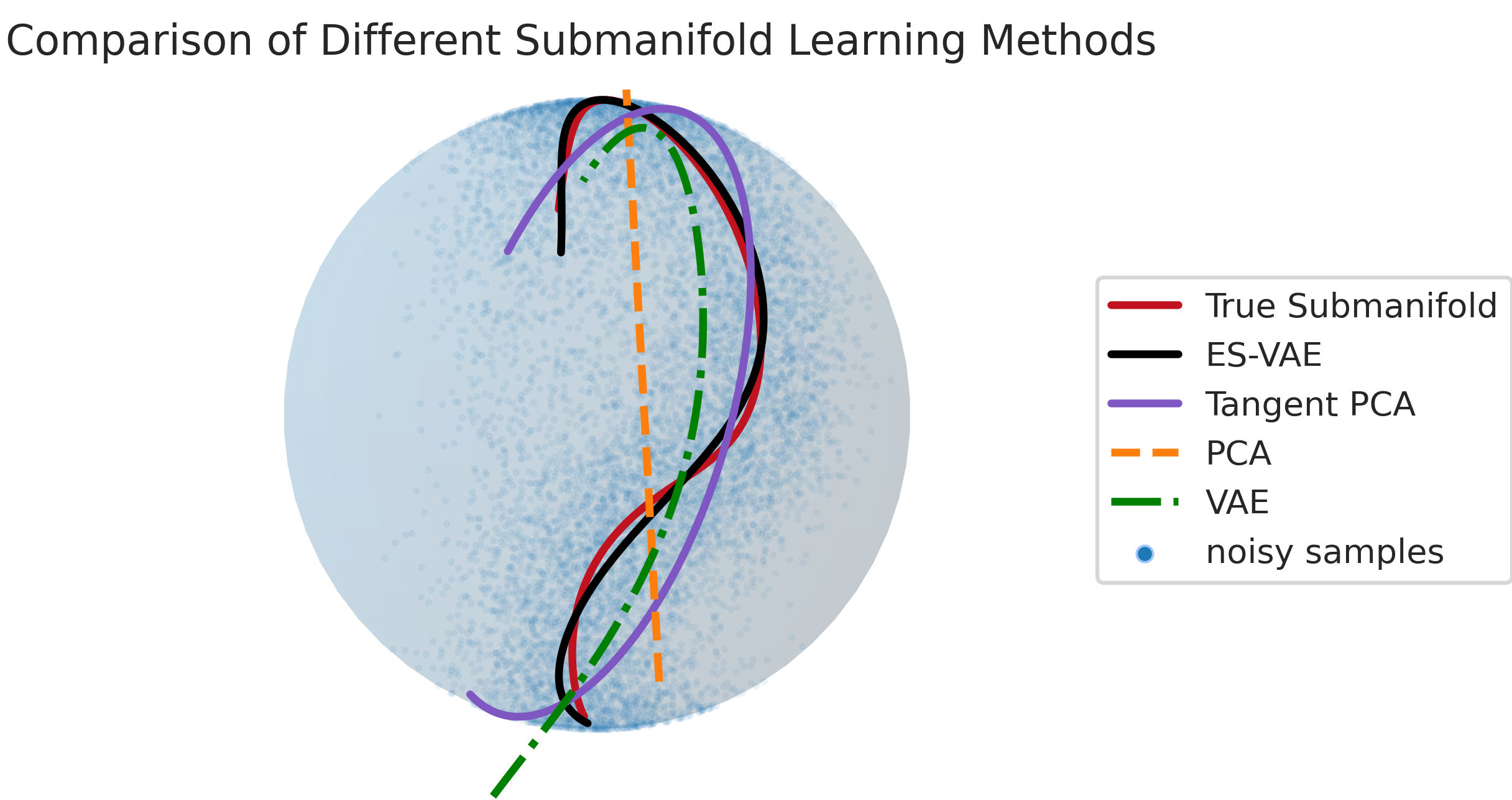}
    \caption{Comparison of submanifold learning methods on synthetic data on $\mathbb{S}^2$. Blue points denote noisy samples generated around a true nonlinear submanifold (red) on the sphere. The black curve shows the submanifold learned by ES-VAE. The purple curve shows the Tangent PCA reconstruction projected back onto the sphere. The dashed orange and dash-dotted green curves correspond to standard Euclidean PCA and Euclidean VAE, respectively. ES-VAE most accurately recovers the data’s underlying non-geodesic structure. In contrast, PCA is limited by its linearity, the VAE fails to capture the spherical geometry in the lower region of the sphere, and Tangent PCA is restricted to geodesic trajectories.}
    \label{fig:simulation}
\end{figure}

We first demonstrate the merit of our approach using synthetic data on a unit sphere $\mathbb{S}^2$. In this experiment, we generate points along a non-geodesic curve -- a one-dimensional submanifold -- on $\mathbb{S}^2$ and used each method to recover that curve (see Figure~\ref{fig:simulation}). PCA and classical VAE both work in the ambient Euclidean space. PCA ignores the curvature entirely, while the VAE cannot follow the spherical geometry in the lower portion. Tangent PCA can only follow geodesic paths because it is linear at the base point. ES-VAE better tracks the non-geodesic structure of the data and gives more accurate approximation of the true submanifold.

\subsection{Regression Performance on Stroke Dataset}
\label{sec:regression}
Next, we present results on clinical gait stroke dataset. Table~\ref{tab:poma_comparison} compares performances of all methods on POMA regression. It can be seen that ES-VAE outperforms others with $R^2 = 0.74$, RMSE $= 2.82$, and Pearson $r = 0.86$ which is clearly better than best raw-skeleton deep learning model (LSTM, $R^2 = 0.64$, RMSE $= 3.31$), and joint angle PCA ($R^2 = 0.44$). Even relative to Tangent PCA ($R^2 = 0.70$, RMSE $= 3.03$), the gap shows the value of the nonlinear VAE mapping over a linear projection in the tangent space. Additional results of classification on stroke dataset is given in Appendix Section~\ref{sec:classification_stroke}.


\begin{table*}[!t]
\centering
\setlength{\tabcolsep}{5pt}
\renewcommand{\arraystretch}{1.08}
\caption{Regression performance for POMA score prediction. Metrics are pooled out-of-fold predictions under subject-wise cross-validation. 95\% confidence intervals are obtained via subject-level bootstrap resampling.}
\label{tab:poma_comparison}
\begin{tabular}{@{}llccc@{}}
\toprule
\makecell[l]{\textbf{Input}\\\textbf{Representation}}
& \textbf{Method}
& \textbf{RMSE}
& $\bm{R^2}$
& \textbf{Pearson $r$} \\
\midrule

\multirow{8}{*}{Raw Skeleton}
& TCN
& \estci{3.82}{3.21}{4.30}
& \estci{0.52}{0.30}{0.67}
& \estci{0.79}{0.73}{0.85} \\
& LSTM
& \estci{3.31}{2.69}{3.79}
& \estci{0.64}{0.52}{0.74}
& \estci{0.81}{0.74}{0.87} \\
& Transformer
& \estci{3.93}{3.19}{4.50}
& \estci{0.50}{0.29}{0.66}
& \estci{0.76}{0.69}{0.84} \\
& ST-GCN
& \estci{3.88}{3.08}{4.49}
& \estci{0.51}{0.34}{0.66}
& \estci{0.72}{0.64}{0.82} \\
& Hyper-GCN
& \estci{3.57}{2.90}{4.07}
& \estci{0.58}{0.48}{0.68}
& \estci{0.77}{0.71}{0.83} \\
& PCA + $k$-NN
& \estci{4.46}{3.91}{4.95}
& \estci{0.35}{0.23}{0.44}
& \estci{0.59}{0.49}{0.68} \\
& VAE + $k$-NN
& \estci{4.33}{3.77}{4.83}
& \estci{0.39}{0.26}{0.48}
& \estci{0.62}{0.52}{0.71} \\

\midrule

\multirow{2}{*}{Joint Angle}
& PCA + $k$-NN
& \estci{4.13}{3.34}{4.72}
& \estci{0.44}{0.34}{0.55}
& \estci{0.76}{0.67}{0.85} \\
& VAE + $k$-NN
& \estci{3.71}{2.87}{4.32}
& \estci{0.55}{0.45}{0.67}
& \estci{0.76}{0.68}{0.84} \\

\midrule

\multirow{2}{*}{Tangent Vector}
& Tangent PCA + $k$-NN
& \estci{3.03}{2.39}{3.48}
& \estci{0.70}{0.59}{0.80}
& \estci{0.84}{0.78}{0.90} \\
& \textbf{ES-VAE + $\bm{k}$-NN (proposed)}
& \bestci{2.82}{2.29}{3.21}
& \bestci{0.74}{0.66}{0.82}
& \bestci{0.86}{0.82}{0.91} \\

\bottomrule
\end{tabular}
\end{table*}

\subsection{Classification Performance on NTU-60 Action Recognition Dataset}
\label{sec:classification_ntu}

Table~\ref{tab:ntu_clf_xsub} reports activity classification results on the NTU-60 dataset in cross-view setup. Here, ES-VAE + k-NN reaches 94.83\% top-1 accuracy against 95.73\% of Hyper-GCN (the strongest full-capacity baseline). Although performance is slightly lower, our model does this without any rotation augmentation, because view invariance is a property of the representation rather than something the network has to learn from augmented examples. However, Hyper-GCN required rotation augmentation in our experiments and is primarily a discriminative classifier and unlike ES-VAE, it is not designed to provide an explicitly interpretable generative latent space. Therefore, ES-VAE is competitive with strong full-capacity graph models on NTU-60 X-View, while providing an explicitly invariant and interpretable latent representation. Finally, ES-VAE achieves the performance with a smaller number of parameters (0.82 M of ES-VAE vs 1.1 M of Hyper-GCN).

\begin{table*}[!t]
\centering
\setlength{\tabcolsep}{5pt}
\renewcommand{\arraystretch}{1.08}
\caption{NTU-60 activity classification results in cross-view setup.}
\label{tab:ntu_clf_xsub}
\begin{tabular}{@{}llcc@{}}
\toprule
\makecell[l]{\textbf{Input}\\\textbf{Representation}}
& \textbf{Method}
& \textbf{Parameters}
& \textbf{Top-1 Accuracy (\%)} \\
\midrule

\multirow{6}{*}{Raw Skeleton}
& TCN
& 1.43 M
& 88.73 \\
& LSTM
& 15.86 M
& 87.63 \\
& Transformer
& 4.92 M
& 82.33 \\
& ST-GCN
& 3.10 M
& 90.10 \\
& Hyper-GCN
& 1.10 M
& 95.73 \\
& PCA + $k$-NN
& N/A
& 52.14 \\
& VAE + $k$-NN
& 0.82 M
& 56.58 \\

\midrule

\multirow{2}{*}{Tangent Vector}
& Tangent PCA + $k$-NN
& N/A
& 64.39 \\
& \textbf{ES-VAE + $\bm{k}$-NN (proposed)}
& \textbf{0.82 M}
& \textbf{94.83} \\

\bottomrule
\end{tabular}
\end{table*}

\subsection{Modes of Variation of Latent Variables in Stroke Dataset}
\label{sec:modes}

\begin{figure}[!t]
    \centering
    \includegraphics[width=0.7\textwidth]{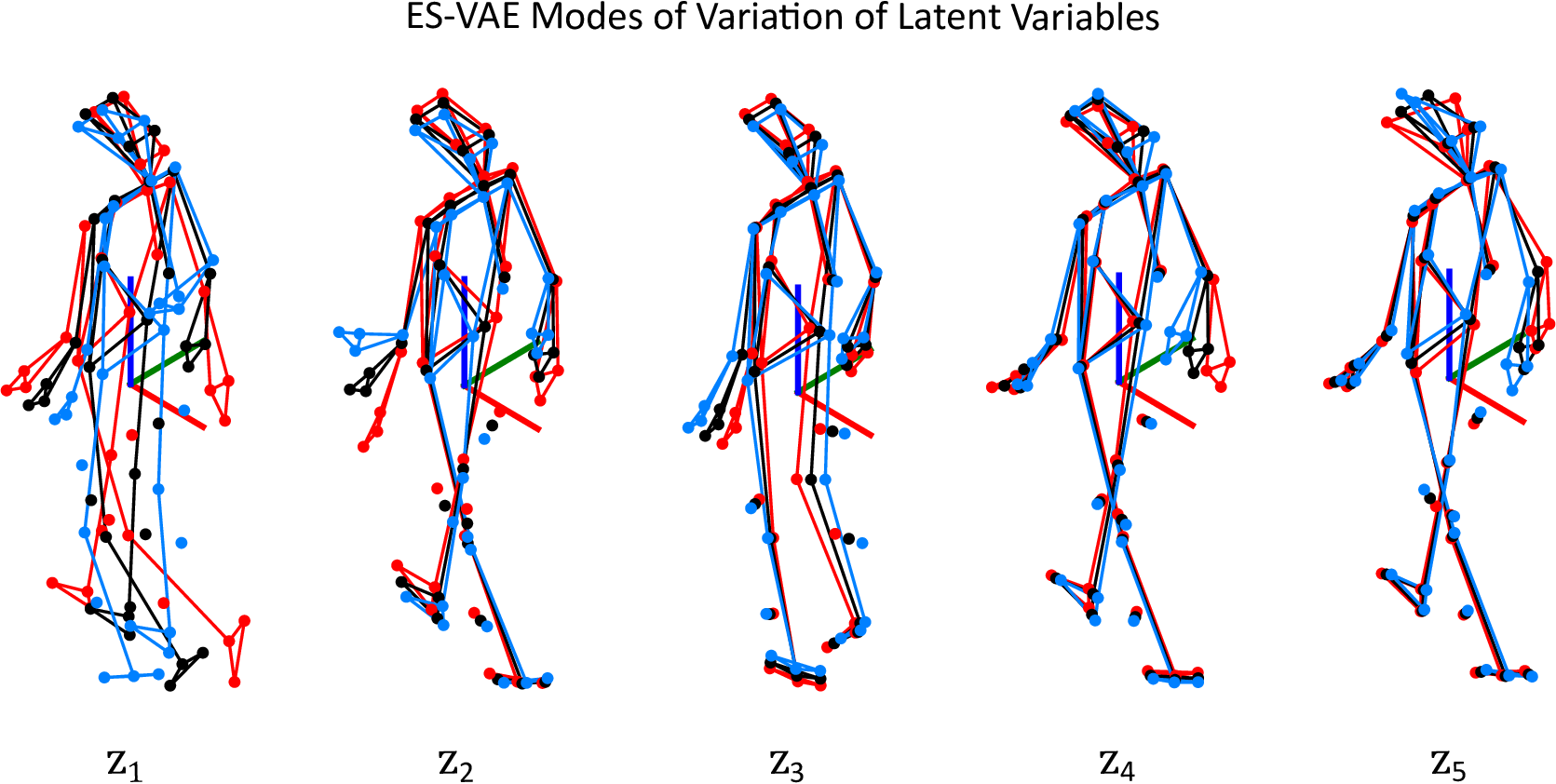}
    \caption{ES-VAE modes of variation: $z_1$ (short stride and stiffer limbs), $z_2$ (left arm variability), $z_3$ (left arm and right knee variability), $z_4$ (right arm variability), $z_5$ (subtle right elbow variability). Black: mean; red/blue: $\pm 3$ traversals.}
    \label{fig:ktrsv_modes}
\end{figure}

Figure~\ref{fig:ktrsv_modes} shows what the first five latent dimensions ($z_1$ through $z_5$) actually encode, sorted by decreasing variance. The mean skeleton is drawn in black and the red/blue overlays show $\pm 3$ unit traversals along each latent dimension while keeping other dimensions fixed. $z_1$ picks up the biggest source of variation: shorter stride length and stiffer limbs, both hallmarks of hemiplegic gait~\citep{li2023stiff}. $z_2$ is about left arm variability, $z_3$ combines left arm and right knee variability, $z_4$ tracks right arm variability, and $z_5$ captures subtle elbow movement differences.

\subsection{Correlation with Clinical Scores in Stroke Dataset}
\label{sec:clinical_corr}

\begin{figure}[!t]
    \centering
    \includegraphics[width=0.75\textwidth]{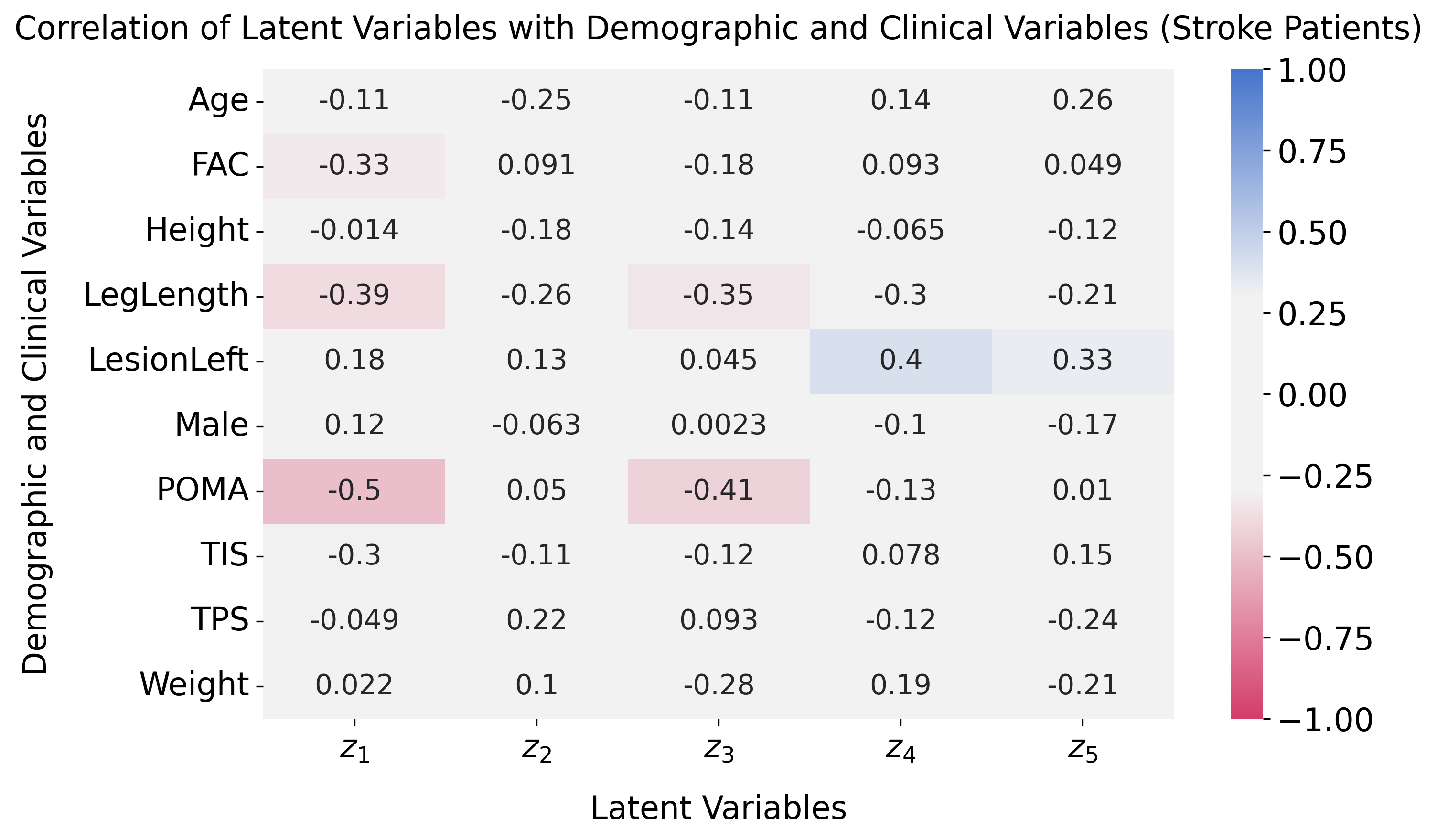}
    \caption{Correlation of demographic and clinical variables with the first five ES-VAE latent dimensions. $z_1$ and $z_3$ correlate negatively with POMA, and $z_4$ correlates with lesion laterality.}
    \label{fig:z_correlation}
\end{figure}

Figure~\ref{fig:z_correlation} shows correlations between the first five ES-VAE latent dimensions and clinical/demographic variables. Both $z_1$ and $z_3$ correlate negatively with POMA ($r = -0.50$ and $r = -0.41$), meaning that participants who score higher on $z_1$ and $z_3$ (shorter stride, stiffer limbs) also score lower on POMA. $z_4$ correlates positively with LesionLeft, which shows that it encodes left side disability.

\subsection{Ablation Studies}
\label{sec:ablation}

We did different ablation studies to understand the effect of different components of our modeling approach. The ablation studies are described in detail in Appendix Section~\ref{sec:appendix_ablation}.

\section{Discussion}
\label{sec:discussion}

Across both datasets, the dominant pattern is that geometry-aware preprocessing matters more than raw model capacity. Once translation, rotation, scale are removed in Kendall shape space and rate variability removed using TSRVF framework, even simple tangent-space baselines become much stronger. ES-VAE improves further by replacing a linear projection with a nonlinear generative model. The analyses show that this gain is not merely predictive: the learned latent dimensions remain interpretable in terms of stride length, limb stiffness, and lateralized arm motion.

The same representation also generalizes across tasks. On stroke gait, it supports clinically meaningful regression and laterality classification; on NTU-60, it improves recognition of actions whose discriminative signal lies in motion shape rather than static pose. This suggests that the main benefit of ES-VAE is structural, namely that it allocates model capacity to shape dynamics after nuisance variability has been removed geometrically.

There are a few clear limitations. The stroke dataset has only $N = 155$ participants and comes from a single center or lab, so we do not yet know how well the approach generalizes to other populations or motion capture setups. The class imbalance (111 healthy vs.\ 14 left stroke vs.\ 30 right stroke) makes laterality classification hard, and the confidence intervals for left-stroke classification are wide. We also kept the encoder-decoder deliberately simple for interpretability. A more expressive architecture (convolutional, attention-based) might push performance further, though probably at the expense of interpretability. Finally, we only use kinematic data. Adding electromyography (EMG) signals could give complementary information about muscle activation patterns.

In future, we plan to scale to larger multi-center datasets, use the model for longitudinal tracking of rehabilitation progress, explore multimodal models incorporating EMG, and conditional VAEs that incorporate participant demographics for personalized activity and gait assessment.

\section{Conclusion}
\label{sec:conclusion}

We presented ES-VAE, a Riemannian VAE that works on skeleton trajectories in Kendall shape space with transported SRVF temporal alignment, making the representation invariant to translations, rotations, scaling, and activity speed all at once. On a stroke dataset, the model learns a compact latent space where individual dimensions map onto recognizable gait features like stride length reduction, limb stiffness, and lateralized arm variability. It predicts POMA scores with $R^2 = 0.74$ and classifies 60 different activities with top-1 accuracy $= 94.83\%$, outperforming most other methods. The results show the advantage of removing geometric noise early using shape-space embedding than to expect a deep network to learn those invariances by itself. Additionally, the nonlinear mapping of a ES-VAE gives a more useful decomposition of gait and action variability than linear tangent PCA. These results make ES-VAE a practical option for interpretable, automated skeleton pose trajectory analysis in stroke rehabilitation and activity recognition.


\bibliographystyle{plainnat}
\bibliography{references}

\appendix

\section{Additional Related Literature}
\label{sec:related_works_appendix}
\noindent {\bf Variational Autoencoders}:
The variational autoencoder~\citep{kingma2013auto} jointly trains an encoder that approximates the posterior and a decoder that reconstructs the data, optimizing the evidence lower bound (ELBO). Kingma and Welling showed that the resulting latent dimensions can pick up meaningful factors like facial orientation and expression. \citet{lopez2018deep} applied the same idea to single-cell transcriptomics (scVI) and found that VAE codes outperform PCA on clustering, imputation, and differential expression. The nonlinear mapping evidently captures structure that linear methods miss, which is what motivates our extension to shape-space data. \citet{chadebec2022data} proposed a geometry-aware VAE whose latent space is modeled as a Riemannian manifold, enabling it to generate meaningful samples from small datasets. Their method was useful for data augmentation in a medical imaging classification task. \citet{shamsolmoali2023vtae} introduced a variational spatial-transformer autoencoder that minimizes geodesic distance on Riemannian manifold and it uses geodesic interpolation for better latent representation, reconstruction, image interpolation. \citet{kalatzis2020variational} argued that the standard Gaussian prior and Euclidean assumptions in VAEs can lead to poor performance, and proposed a Riemannian Brownian motion prior that provides the latent space with a Riemannian structure, significantly improving the model's representational capacity. In most of these works, the latent space is modeled as a Riemannian manifold. In contrast, our approach represents the skeleton input in Kendall shape space, which is itself a Riemannian manifold.

\noindent {\bf Stroke Gait Assessment}:
On the clinical side, dynamic time warping (DTW) has been used traditionally for comparing gait patterns. \citet{tormene2009matching} matched incomplete rehabilitation time series with DTW, \citet{lee2019application} validated it for gait pattern similarity, and \citet{Zhang2016Objective} used DTW distance to measure upper-limb mobility post-stroke. \citet{eichler20183d} built a 3D motion capture system that registers skeletons rotationally via singular value decomposition (SVD) for Fugl-Meyer assessment. Joint angle PCA~\citep{cho2024stroke} takes a different route, computing principal components of joint angle time series. \citet{li2023stiff} characterized stiff knee gait as a neuromechanical consequence of spastic hemiplegia. The trouble with DTW is that it is not a proper metric and ignores rotation and scaling. Joint angle PCA, on the other hand, throws away the full 3D spatial information. Working in Kendall shape space avoids both problems because it gives a proper Riemannian metric that is invariant to all of these nuisance transformations.

\section{Derivation of the Riemannian ELBO}
\label{app:elbo}

We provide a detailed derivation of the loss function used to train ES-VAE.

\subsection{Evidence Lower Bound}
Starting from Bayes' rule:
\begin{align}
p(z \mid x) = \frac{p(x \mid z)\,p(z)}{p(x)},
\end{align}
we take logarithms and rearrange:
\begin{align}
\ln p(x) = \ln p(x \mid z) - \ln p(z \mid x) + \ln p(z).
\end{align}

Introducing the variational approximation $q(z \mid x)$ by adding and subtracting $\ln q(z \mid x)$, then taking the expectation under $q(z \mid x)$:
\begin{align}
\ln p(x) &= \underbrace{\mathbb{E}_{q(z|x)}[\ln p(x|z)] - \mathrm{KL}(q(z|x) \| p(z))}_{\text{ELBO}} \notag \\
&\quad + \underbrace{\mathrm{KL}(q(z|x) \| p(z|x))}_{\geq 0}.
\end{align}

Since the KL divergence is non-negative, the ELBO is a lower bound on $\ln p(x)$, and maximizing the ELBO is equivalent to minimizing the KL divergence between $q$ and the true posterior.

\subsection{KL Divergence for Gaussian Prior}
For the standard normal prior $p(z) = \mathcal{N}(\mathbf{0}, \mathbf{I})$ and a diagonal Gaussian encoder $q_\phi(z \mid v) = \mathcal{N}(\mu(v), \mathrm{diag}(\sigma^2(v)))$ with $L$ latent dimensions:
\begin{align}
\mathrm{KL}_n &= \frac{1}{2}\sum_{j=1}^{L}\left(\sigma_j^2(v_n) + \mu_j^2(v_n) - 1 - \log \sigma_j^2(v_n)\right).
\end{align}

For a mini-batch of $N$ samples: $\mathrm{KL} = \frac{1}{N}\sum_{n=1}^{N}\mathrm{KL}_n$.

\subsection{Riemannian Reconstruction Loss}
For data on a Riemannian manifold $\mathcal{M}$, the reconstruction term replaces Euclidean distance with the squared geodesic distance. For a trajectory with $T$ time steps:
\begin{align}
\mathrm{Recon}_n = \sum_{t=1}^{T} d_{\Sigma_m^k}\!\left(\tilde{\beta}_n(t),\, \hat{\beta}_n(t)\right)^2,
\end{align}
where $d_{\Sigma_m^k}$ is the geodesic distance on the shape manifold, $\tilde{\beta}_n(t)$ is the input, and $\hat{\beta}_n(t) = \mathrm{Exp}_\mu(f_\theta(z_n))(t)$ is the reconstruction.

\subsection{Total Loss}
The total loss to minimize (negative ELBO) over a mini-batch:
\begin{align}
\mathcal{L} &= \frac{1}{N}\sum_{n=1}^{N}\left(\mathrm{Recon}_n + \beta_{\text{KL}} \cdot \mathrm{KL}_n\right) \notag \\
&= \frac{1}{N}\sum_{n=1}^{N}\Bigg(\sum_{t=1}^{T} d_{\Sigma_m^k}(\tilde{\beta}_n(t), \hat{\beta}_n(t))^2 \notag
+ \frac{\beta_{\text{KL}}}{2}\sum_{j=1}^{L}\left(\sigma_j^2 + \mu_j^2 - 1 - \log \sigma_j^2\right)\Bigg),
\end{align}
where $\beta_{\text{KL}}$ weights the KL term relative to reconstruction.

\section{Dataset Details}
\label{sec:appendix_dataset}
\subsection{Stroke Gait Dataset}
\label{sec:dataset_stroke}
We use the publicly available full-body motion capture gait dataset of~\citet{van2023full}, which contains 138 able-bodied adults and 50 stroke survivors. After excluding participants with incomplete recordings, our analysis includes $N = 155$ participants: 111 healthy controls and 44 stroke patients (14 with left-sided and 30 with right-sided hemiplegia). Each participant performed overground walking at a self-selected speed while $k = 32$ reflective markers were tracked by an optical motion capture system, recording $m = 3$ spatial coordinates per marker. Gait cycles were segmented and resampled to $T = 200$ time steps, yielding trajectories of dimension $k \times m \times T = 32 \times 3 \times 200 = 19{,}200$ per participant. Clinical assessments include the Tinetti Performance Oriented Mobility Assessment (POMA), Functional Ambulation Category (FAC), and lesion laterality (left or right hemiplegia).

\subsection{NTU-60 Action Recognition Subset}
\label{sec:dataset_ntu}
To test whether the manifold prior generalizes beyond clinical gait data, we evaluate the same Kendall shape-space pipeline on the full NTU RGB+D (NTU-60) dataset~\citep{shahroudy2016ntu}. We used 56,880 sequences, all 60 classes, without any curation. We chose X-View deliberately. It is a camera-viewpoint generalisation test, which is exactly the nuisance factor Kendall shape space removes by construction rather than by augmentation. Each trial is a sequence of $k = 25$ skeleton joints with $m = 3$ spatial coordinates, linearly interpolated to $T = 100$ time steps, yielding tensors of dimension $25 \times 3 \times 100 = 7{,}500$ per trial.

\section{Additional Implementation Details}\label{sec:addtional_implementation}
\subsection{Network Architecture} 
The encoder is a linear layer mapping the $D$-dimensional tangent vector to a hidden layer of size $H$, followed by $\tanh$ activation and dropout ($p = 0.1$), then two parallel linear heads producing the mean $\mu_\phi \in \mathbb{R}^L$ and log-variance $\log \sigma^2_\phi \in \mathbb{R}^L$, where $L$ is the latent dimension. The decoder consists of two linear layers ($L \to H' \to D$) with $\tanh$ activation in between. Latent samples are obtained via the reparameterization trick: $z = \mu_\phi + \sigma_\phi \odot \epsilon$, $\epsilon \sim \mathcal{N}(\mathbf{0}, \mathbf{I})$. The dataset-specific widths are:

\textbf{Stroke dataset.} Tangent vector dimension $D = 32 \times 3 \times 200 = 19{,}200$, hidden width $H = 128$, decoder hidden $H' = 16$, latent dimension $L = 38$. The hyperparameters $(L, H)$ were selected by a grid search using ELBO loss on the validation folds where $L \in \{16, 24, 32, 38, 48, 64\}$, and $H \in \{8, 16, 32, 64, 128, 256\}$.

\textbf{NTU-60 dataset.} Tangent vector dimension $D = 25 \times 3 \times 100 = 7500$, hidden width $H = 768$, decoder hidden $H' = H$, latent dimension $L = 48$. The hyperparameters $(L, H)$ were selected by a grid search using ELBO loss on the validation folds where $L \in \{16, 24, 32, 48, 64\}$, and $H \in \{512, 768\}$.

\subsection{Training}
Models are trained with the Adam optimizer at learning rate $10^{-3}$ on the ELBO with squared geodesic reconstruction loss and a Gaussian KL prior weighted by $\beta_{\text{kl}}$. Full budget hyperparameter search was done for vanilla VAE similar to ES-VAE to achieve highest possible performance. The KL weight $\beta_{\text{kl}}$ was selected using grid search: $\beta_{\text{kl}} \in \{2^{-5}, 2^{-4}, 2^{-3}, 2^{-2}, 2^{-1}\}$ for the stroke dataset and $\beta_{\text{kl}} \in \{10^{-5}, 10^{-4}, 10^{-3}, 10^{-2}, 10^{-1}\}$ for the NTU-60 dataset. For NTU-60, the final ES-VAE used 150 epochs, batch size 64, and $\beta=10^{-4}$. For the stroke dataset, the final ES-VAE used 100 epochs, full-batch training, and $\beta_{\text{kl}}=2^{-3}$.

\subsection{Inference}
After training, latent dimensions are reordered by decreasing variance across the dataset so that $z_1$ captures the largest source of variation. Downstream prediction uses $k$-nearest-neighbour retrieval in the latent space with inverse-distance weighting, using the scikit-learn default of $k = 5$ on both datasets. Test sequences are encoded with the trained encoder and matched against the training-fold latent codes.

\subsection{Evaluation Protocol}
\label{sec:eval}
\textbf{Stroke dataset.} All metrics are computed from pooled out-of-fold test predictions under subject-wise cross-validation with 30 folds: 28 folds arranged as two rounds of 14 folds, each with 5 validation and 5 test participants; one fold with 10 validation and 5 test participants; and one fold with 5 validation and 10 test participants. Standardization and Fr\'{e}chet mean fitted on the training set of each fold. Confidence intervals are obtained via subject-level bootstrap resampling (2000 replicates).

\textbf{NTU-60 dataset.} We evaluate on the full NTU RGB+D 60 dataset using the standard cross-view (X-View) protocol across all 60 action classes. Sequences captured by cameras 2 and 3 are used for training, while those captured by camera 1 are reserved for testing. Standardization parameters and the Fr\'{e}chet mean are estimated exclusively from the training set and held fixed during testing. We report top-1 classification accuracy (\%) on the held-out test set.

\subsection{Baseline Architectures and Compute Parity} \label{sec:baseline-details}

\textbf{NTU-60 sequence baselines.} All four models are trained for $150$ epochs at learning rate $2\!\times\!10^{-3}$, weight decay $10^{-4}$, batch size $64$, and label smoothing $0.05$ on cross-entropy. (i) \emph{TCN}: two residual temporal blocks with $16$ channels each, kernel size $3$, dilations $\{1, 2\}$, BatchNorm + GELU + weight-norm convolutions, dropout $0.40$; the head concatenates global-mean and global-max pooled features and applies a 2-layer MLP of width $24$ to the $60$-class logits. (ii) \emph{LSTM}: a single unidirectional layer with hidden size $16$, dropout $0.40$, mean+max pooled over time, and the same 2-layer head. (iii) \emph{Transformer}: a $1$-layer pre-norm encoder with $d_\text{model}=24$, $2$ heads, $d_\text{ff}=48$, dropout $0.30$, sinusoidal positional encoding, and mean pooling. (iv) \emph{ST-GCN}: a $2$-block spatio-temporal GCN with channel widths $(16, 32)$, temporal kernel $9$, dropout $0.30$, NTU-25 kinematic adjacency with symmetric normalization, and global average pooling over time and joints. Total trainable parameters are well under $50$k for every sequence baseline, deliberately keeping their capacity comparable to the ES-VAE encoder. (v) \emph{Hyper-GCN}: this baseline is run from unmodified architecture code copied from the official release. It uses the published variant with $25$ joints, $\texttt{hyper\_joints}=3$, the NTU-RGB+D graph in \texttt{virtual\_ensemble} mode, SGD with Nesterov momentum $0.9$, base learning rate $0.05$, weight decay $4\!\times\!10^{-4}$, label smoothing $0.1$, the official $5$-epoch warm-up + step decay at epochs $[110, 120]$, the divergence loss on virtual hyper-joints, and a clip window of $64$ frames. We deliberately did not retune any of these official hyperparameters for the small-data regime: the goal of including them is to expose, not erase, the overfitting characteristic of full-capacity NTU graph backbones.

\textbf{Stroke gait baselines.} Each subject is encoded as a tangent-vector array of shape $(32, 3, 200)$ (or the matched raw-marker tensor of the same shape). All sequence baselines slide a window of size $100$ with stride $25$ across the temporal axis ($5$ windows per trial), train on per-window targets with subject-level WeightedRandomSampler ($1/\sqrt{n_c}$ class weights) for the 3-class classifier, and aggregate window predictions per subject by median (regression) or softmax averaging followed by argmax (classification). All four sequence models share Adam with learning rate $5\!\times\!10^{-4}$, weight decay $10^{-4}$, batch size $32$, cosine annealing, gradient clipping $1.0$, SmoothL1 loss for regression, and cross-entropy for classification. The per-model widths are: (i) \emph{TCN}: three temporal blocks, channels $(64, 64, 96)$ for regression and $(32, 32)$ for classification, kernel size $5$, dropout $0.25$/$0.35$, head hidden width $64$/$32$. (ii) \emph{LSTM}: $1$ layer, hidden $64$ (regression) or $32$ (classification), dropout $0.15$, unidirectional, last-step + projection head. (iii) \emph{Transformer}: $1$ encoder layer, $d_\text{model}=24$, $d_\text{ff}=64$, $2$ heads, dropout $0.10$ (regression); $d_\text{model}=12$, $d_\text{ff}=32$, $2$ heads, dropout $0.20$ (classification). (iv) \emph{ST-GCN}: a $32$-node skeleton graph with the symmetric-normalized adjacency, two ST blocks with channel widths $(32, 64)$ for regression and $(16, 32)$ for classification, temporal kernel $7$, dropout $0.20$/$0.30$. The default training horizons for the LSTM/Transformer/ST-GCN classifier and regressor are $6$ and $8$ epochs respectively (the small-window-multiplied effective batch count is large), while TCN trains substantially longer because its much higher capacity needs more updates to reach plateau on the same $\approx\!145$-subject training fold. (v) \emph{Hyper-GCN}: it reuses the unmodified official architectures with the same regression/classification head adaptations as in the NTU runs and identical optimizer/schedule defaults.

\textbf{Compute parity statement.} For every dataset and every CV fold, all baselines and ES-VAE see (a) identical training/test indices, (b) identical preprocessing (per-channel $z$-score, identical windowing where applicable), (c) the same deterministic seed, and (d) the same evaluation pipeline. All the models were tuned to achieve their best performance.

\subsection{Compute Resources}
\label{sec:compute_resource}
All experiments were run on a single workstation with two Intel Xeon Silver 4210R CPUs (20 cores / 40 threads, 2.40 GHz), 128 GB RAM, and one NVIDIA RTX A5000 GPU (24 GB VRAM), running CUDA 11.8 with PyTorch 2.5.1. We used FDASRSF~\citep{tucker2013generative} and Geomstats~\citep{JMLR:v21:19-027} packages for shape alignment. No multi-GPU or cluster compute was used.

\begin{table*}[!b]
\centering
\setlength{\tabcolsep}{5pt}
\renewcommand{\arraystretch}{1.08}
\caption{Three-class classification performance for Healthy vs.\ Left Stroke vs.\ Right Stroke. Metrics are macro-averaged F1-score, precision, and recall under subject-wise cross-validation. Values are point estimates with raised 95\% bootstrap confidence intervals in parentheses.}
\label{tab:clf_comparison}
\begin{tabular}{@{}llccc@{}}
\toprule
\makecell[l]{\textbf{Input}\\\textbf{Representation}}
& \textbf{Method}
& \textbf{Macro F1}
& \textbf{Macro Precision}
& \textbf{Macro Recall} \\
\midrule

\multirow{8}{*}{Raw Skeleton}
& TCN
& \estci{0.64}{0.53}{0.72}
& \estci{0.70}{0.53}{0.87}
& \estci{0.62}{0.54}{0.69} \\
& LSTM
& \estci{0.56}{0.48}{0.63}
& \estci{0.57}{0.47}{0.67}
& \estci{0.56}{0.49}{0.62} \\
& Transformer
& \estci{0.63}{0.54}{0.70}
& \estci{0.63}{0.54}{0.72}
& \estci{0.63}{0.54}{0.72} \\
& ST-GCN
& \estci{0.39}{0.33}{0.44}
& \estci{0.48}{0.38}{0.59}
& \estci{0.40}{0.36}{0.44} \\
& Sparse ST-GCN
& \estci{0.48}{0.40}{0.55}
& \estci{0.75}{0.65}{0.83}
& \estci{0.49}{0.43}{0.56} \\
& Hyper-GCN
& \estci{0.47}{0.41}{0.54}
& \estci{0.48}{0.41}{0.55}
& \estci{0.47}{0.40}{0.54} \\
& PCA + $k$-NN
& \estci{0.55}{0.46}{0.62}
& \estci{0.55}{0.46}{0.64}
& \estci{0.55}{0.47}{0.63} \\
& VAE + $k$-NN
& \estci{0.61}{0.51}{0.69}
& \estci{0.68}{0.51}{0.85}
& \estci{0.58}{0.51}{0.67} \\

\midrule

\multirow{2}{*}{Joint Angle}
& PCA + $k$-NN
& \estci{0.61}{0.50}{0.70}
& \estci{0.78}{0.53}{0.90}
& \estci{0.57}{0.49}{0.64} \\
& VAE + $k$-NN
& \estci{0.69}{0.60}{0.78}
& \estci{0.78}{0.66}{0.89}
& \estci{0.65}{0.57}{0.74} \\

\midrule

\multirow{8}{*}{Tangent Vector}
& TCN
& \estci{0.75}{0.66}{0.83}
& \estci{0.78}{0.68}{0.89}
& \estci{0.74}{0.66}{0.83} \\
& LSTM
& \estci{0.70}{0.60}{0.79}
& \estci{0.81}{0.69}{0.91}
& \estci{0.67}{0.59}{0.75} \\
& Transformer
& \estci{0.63}{0.51}{0.71}
& \estci{0.75}{0.51}{0.87}
& \estci{0.60}{0.52}{0.67} \\
& ST-GCN
& \estci{0.48}{0.42}{0.52}
& \estci{0.50}{0.44}{0.56}
& \estci{0.48}{0.43}{0.53} \\
& Sparse ST-GCN
& \estci{0.53}{0.43}{0.62}
& \estci{0.73}{0.62}{0.85}
& \estci{0.49}{0.44}{0.55} \\
& Hyper-GCN
& \estci{0.56}{0.50}{0.62}
& \estci{0.56}{0.49}{0.63}
& \estci{0.57}{0.51}{0.63} \\
& Tangent PCA + $k$-NN
& \estci{0.79}{0.70}{0.87}
& \estci{0.83}{0.75}{0.92}
& \estci{0.76}{0.68}{0.85} \\
& \textbf{ES-VAE + $\bm{k}$-NN (proposed)}
& \bestci{0.83}{0.75}{0.90}
& \bestci{0.86}{0.79}{0.93}
& \bestci{0.80}{0.73}{0.89} \\

\bottomrule
\end{tabular}
\end{table*}


\section{Additional Results}

\subsection{Mean Shapes in Stroke Dataset}
\label{sec:mean_pairwise}

\begin{figure*}[!t]
    \centering
    \includegraphics[width=0.99\textwidth]{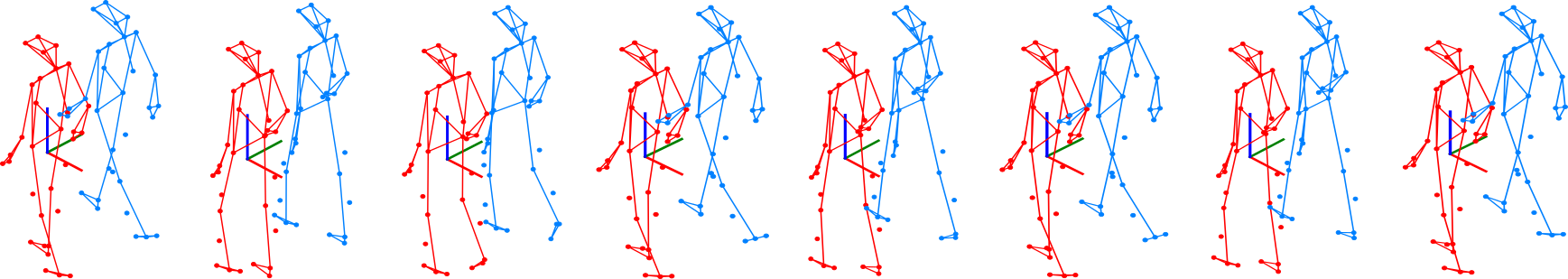}
    \caption{Mean gait cycles from registered trajectories. Stroke patients (red) exhibit hemiplegic dragging, shorter stride length, and stiffer limb motion compared to healthy subjects (blue).}
    \label{fig:means}
\end{figure*}

The mean gait cycles computed from registered trajectories are shown in Figure~\ref{fig:means}. Stroke patients (red) exhibit characteristic hemiplegic dragging, reduced stride length, and stiffer limb motion compared to the fluid, symmetric gait of healthy subjects (blue).

\subsection{Classification Performance on Stroke Dataset}
\label{sec:classification_stroke}

Table~\ref{tab:clf_comparison} gives the three-class classification results (Healthy vs.\ Left Stroke vs.\ Right Stroke). ES-VAE leads with a macro F1 of 0.83. Raw-skeleton deep learning methods lag well behind: LSTM gets 0.56, TCN 0.64, Transformer 0.63. The fact that tangent-vector methods (Tangent PCA 0.79, ES-VAE 0.83) do so much better than raw-skeleton methods really underlines how much removing geometric confounders matters.

Looking at the per-class breakdown in Table~\ref{tab:lesion_classification}, ES-VAE identifies healthy subjects almost perfectly (F1 $= 0.97$, recall $= 1.00$). Separating left from right stroke is harder, but the model still does well: F1 $= 0.72$ for left stroke, F1 $= 0.79$ for right stroke, and 0.92 overall accuracy.

\begin{table}[!t]
\centering
\caption{Per-class classification report for ES-VAE + $k$-NN.}
\label{tab:lesion_classification}
\begin{tabular}{lcccc}
\toprule
\textbf{Class} & \textbf{Precision} & \textbf{Recall} & \textbf{F1}\\
\midrule
Stroke (Left)  & 0.71 & 0.71 & 0.72\\
Stroke (Right)  & 0.91 & 0.70 & 0.79\\
Healthy     & 0.94 & 1.00 & 0.97\\
\midrule
Accuracy        &      &      & 0.92\\
Macro Avg       & 0.86 & 0.80 & 0.83\\
Weighted Avg    & 0.91 & 0.92 & 0.91\\
\bottomrule
\end{tabular}
\end{table}

\subsection{Independence and Separability of Latent Variables in Stroke Dataset}
\label{sec:latent_analysis}

\begin{figure}[!t]
    \centering
    \includegraphics[width=0.9\textwidth]{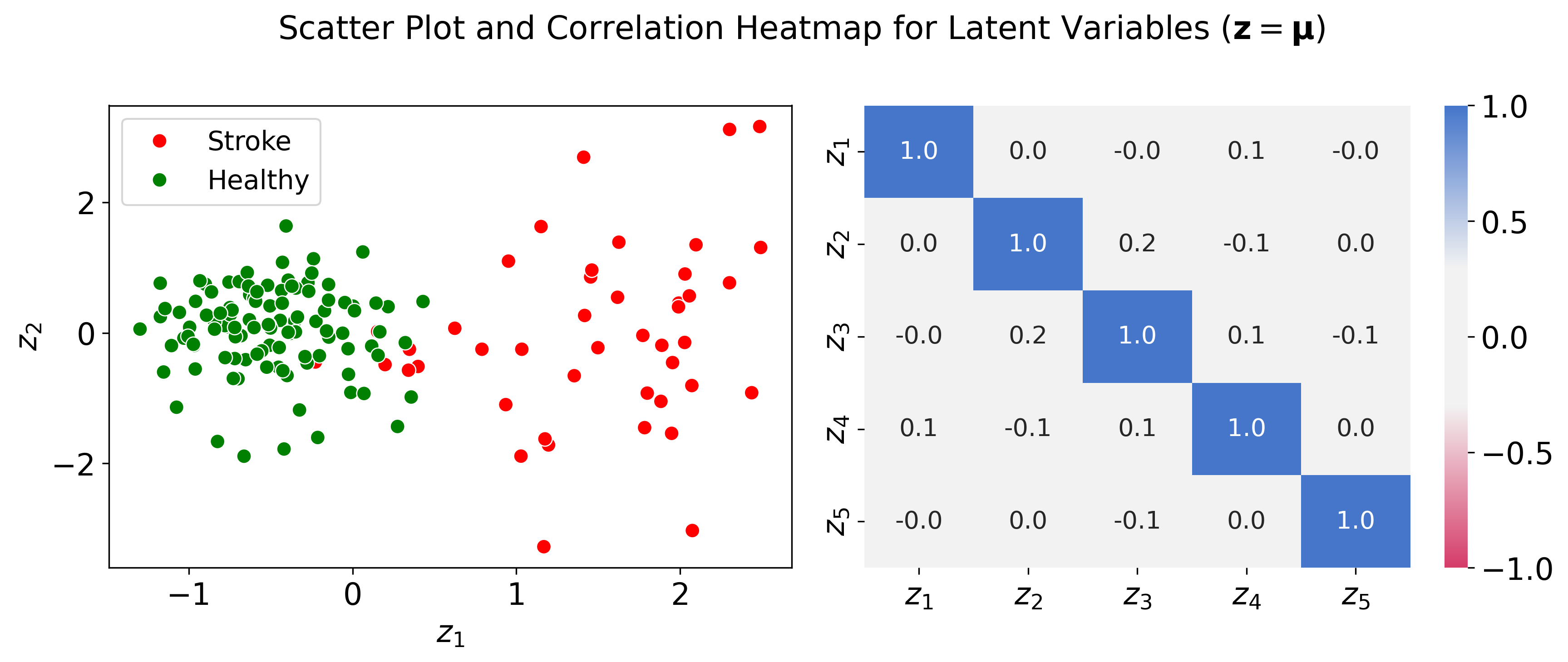}
    \caption{Left: Scatter plot of $z_1$ vs.\ $z_2$ showing separation between stroke (red) and healthy (green) cohorts. Right: Correlation heatmap of the first five latent dimensions, confirming their independence.}
    \label{fig:scatter_heatmap}
\end{figure}

Figure~\ref{fig:scatter_heatmap} shows a scatter plot of $z_1$ versus $z_2$ together with a correlation heatmap of the first five latent dimensions. The heatmap shows that the latent variables are largely uncorrelated, so each dimension really is picking up different aspects of gait. In the scatter plot, healthy participants cluster tightly near the origin while stroke patients spread out much more, which fits with how variable post-stroke gait tends to be.

\begin{figure}[!t]
    \centering
    \includegraphics[width=0.85\textwidth]{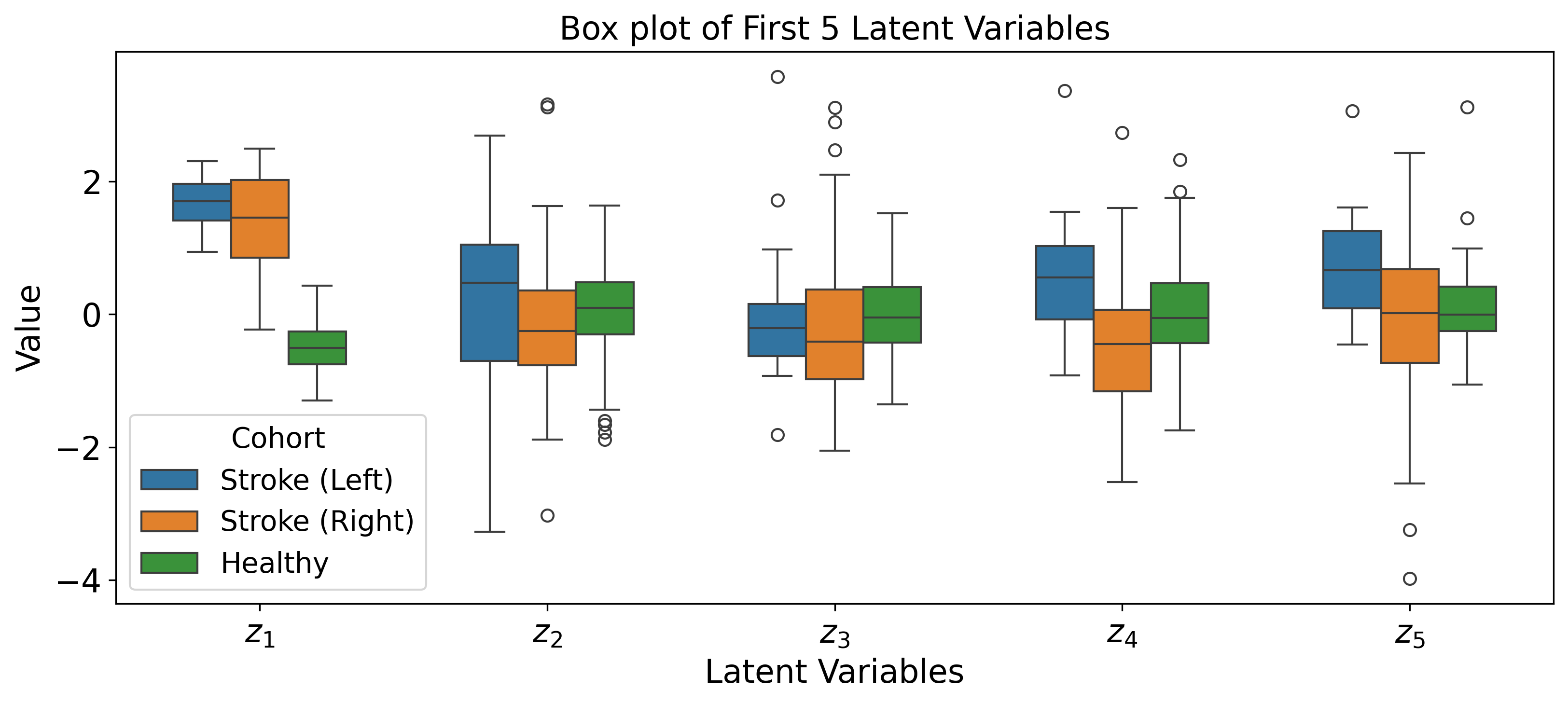}
    \caption{Boxplots of the first five latent dimensions by clinical group. $z_1$ gives the widest separation between Stroke and Healthy, and $z_4$ best separates Left from Right hemiplegia.}
    \label{fig:boxplot}
\end{figure}

Figure~\ref{fig:boxplot} breaks this down further with boxplots by clinical group. $z_1$ gives the widest gap between Stroke and Healthy, as expected from its link to stride length and limb stiffness. $z_4$ is the most useful for telling Left from Right stroke apart, consistent with it encoding unilateral arm variability.

\subsection{Correlation with Clinical Scores and Tangent PCA in Stroke Dataset}
\begin{figure}[!t]
    \centering
    \includegraphics[width=0.82\textwidth]{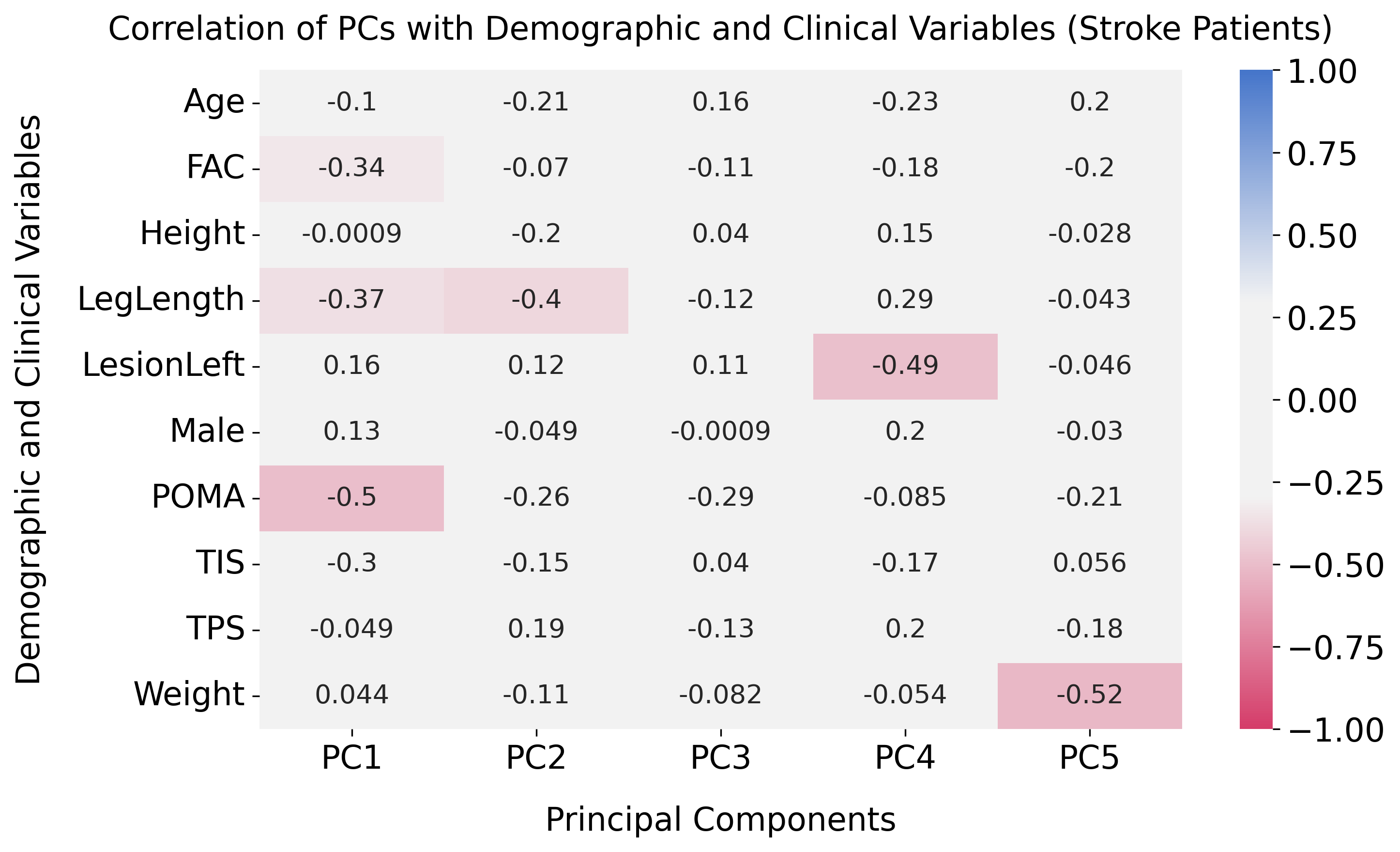}
    \caption{Correlation of demographic and clinical variables with the first five Tangent PCA dimensions.}
    \label{fig:pc_correlation}
\end{figure}

Figure~\ref{fig:pc_correlation} shows that only PC1 reaches $r = 0.50$ with POMA and PC4 shows $-0.49$ with LesionLeft. So ES-VAE spreads the clinical information over more than one latent dimension ($z_1$ and $z_3$ both carry POMA signal), which may give a richer picture than Tangent PCA's single-dimension correlations.

\subsection{Tangent PCA Modes of Variation in Stroke Dataset}
\begin{figure}[]
    \centering
    \includegraphics[width=0.7\textwidth]{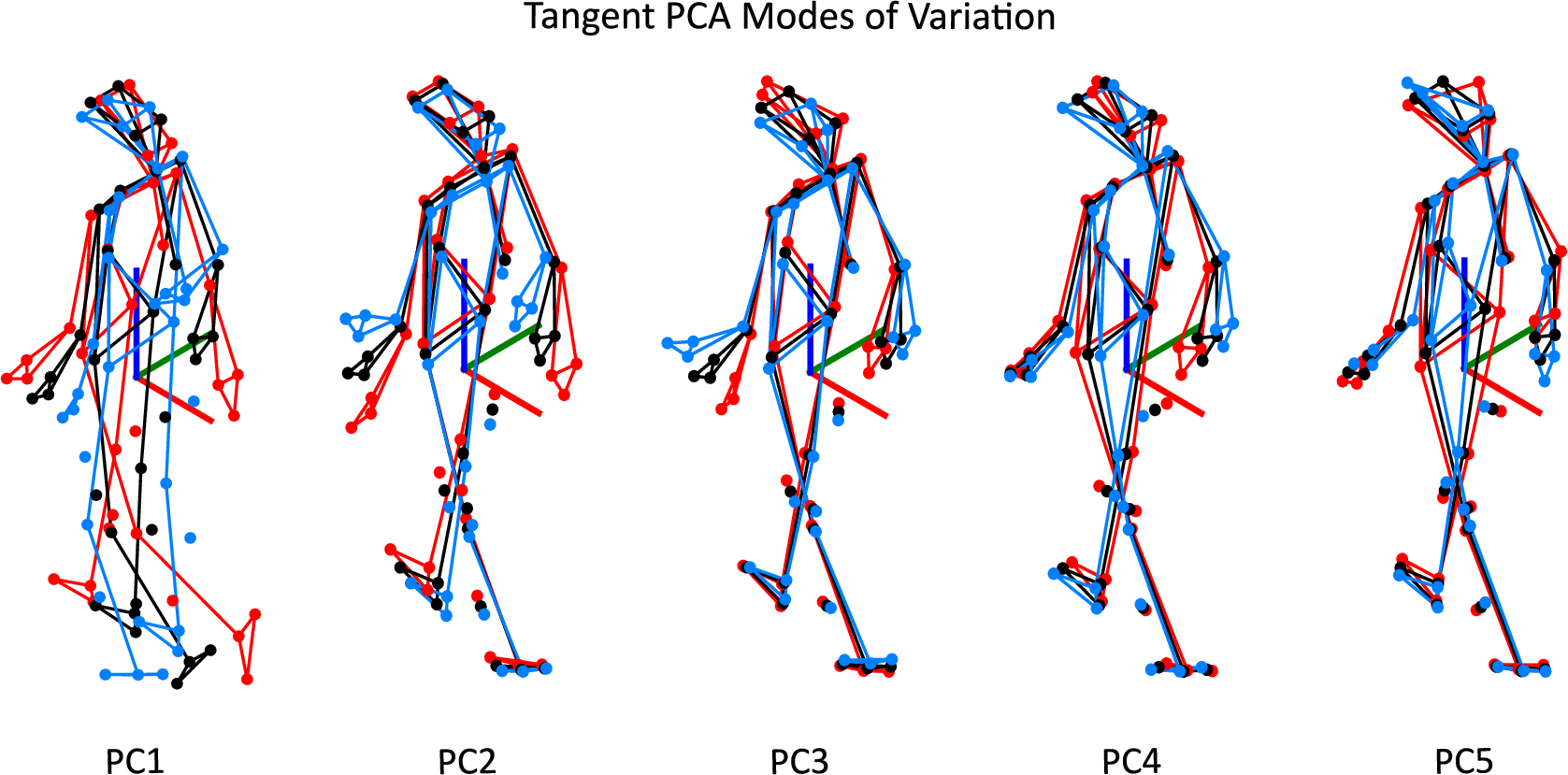}
    \caption{Tangent PCA modes of variation: PC1 (stride/stiffness), PC2 (bilateral arm variability), PC3 (left arm with some right arm), PC4 (right arm with small left hip), PC5 (right hip variability). Black: mean; red/blue: $\pm 3$ traversals.}
    \label{fig:pca_modes}
\end{figure}

Figure~\ref{fig:pca_modes} shows the first five Tangent PCA modes. PC1 looks a lot like $z_1$ (stride/stiffness), but the higher modes diverge. PC2 mixes both arms together, while ES-VAE's $z_2$ isolates the left arm. PC3 in Tangent PCA blends left arm with some right arm variability, whereas ES-VAE's $z_3$ pairs the left arm with the right knee. PC5 picks up right hip variability, while ES-VAE's $z_5$ focuses on the elbow. Overall, ES-VAE seems to decompose gait variability in a way that separates individual limbs more cleanly.

\section{Ablation Studies}
\label{sec:appendix_ablation}

\textbf{Effect of different alignments.}
Table~\ref{tab:alignment_ablation_esvae} shows the benefit of progressively stronger geometric alignment. On NTU-60, Macro F1 improves steadily from 0.57 without alignment to 0.95 under full alignment, while on the stroke dataset RMSE decreases from 5.75 to 2.82. Notably, translation removal alone is insufficient for stroke severity regression, whereas the major gains appear after removing scale and rotation, with the best performance achieved by the fully aligned Kendall + TSRVF representation.

\begin{table}[!t]
\centering
\setlength{\tabcolsep}{5pt}
\renewcommand{\arraystretch}{1.08}
\caption{Alignment ablation with ES-VAE across the NTU activity-recognition dataset and the stroke dataset. Values are point estimates of F1 and RMSE with raised 95\% confidence intervals in parentheses.}
\label{tab:alignment_ablation_esvae}
\begin{tabular}{@{}lcc@{}}
\toprule
\textbf{Alignment stage} & \textbf{NTU-60 Top-1 Accuracy (\%)} & \textbf{Stroke RMSE} \\
\midrule
No alignment
& 0.57
& 5.75 \\
Translation removal (centering)
& 0.74
& 6.01 \\
Translation + scale removal (preshape)
& 0.77
& 5.72 \\
Translation + scale + rotation removal (Kendall)
& 0.84
& 3.53 \\
Full alignment (Kendall + TSRVF)
& 0.95
& 3.21 \\
\bottomrule
\end{tabular}
\end{table}

\textbf{KL weight $\beta_{\text{kl}}$.} We swept $\beta_{\mathrm{kl}}$ over dataset-specific ranges for both the stroke and NTU-60 datasets (Table~\ref{tab:esvae_beta_sweep}). In both cases, intermediate KL weights gave the best trade-off between latent regularization and downstream performance: for the stroke dataset, $\beta_{\mathrm{kl}}=2^{-3}$ achieved the best $R^2$, while for NTU-60, $\beta_{\mathrm{kl}}=10^{-4}$ gave the highest Macro F1. Larger $\beta$ values made the latent space more Gaussian but degraded reconstruction and predictive performance, whereas very small $\beta$ values could lead to weakly regularized or partially collapsed latent dimensions.

\begin{table}[!ht]
\centering
\caption{Effect of the KL-weight $\beta_{\text{kl}}$ on ES-VAE performance for NTU-60 action recognition and the stroke dataset.}
\label{tab:esvae_beta_sweep}
\renewcommand{\arraystretch}{1.25}
\begin{tabular}{l|ccccc|lllll}
\hline
\textbf{Metric}   & \multicolumn{5}{c|}{\textbf{NTU-60 Macro F1}}                  & \multicolumn{5}{c}{\textbf{Stroke $R^2$}}                                                                                                        \\
\hline
$\beta_\text{kl}$ & $10^{-5}$ & $10^{-4}$   & $10^{-3}$ & $10^{-2}$ & $10^{-1}$ & $2^{-5}$  & $2^{-4}$ & $2^{-3}$  & $2^{-2}$  & $2^{-1}$ \\
\hline
Value   & 0.91     & \textbf{0.95} & 0.87     & 0.82     & 0.74     & \multicolumn{1}{c}{0.737} & \multicolumn{1}{c}{0.737} & \multicolumn{1}{c}{\textbf{0.740}} & \multicolumn{1}{c}{0.737} & \multicolumn{1}{c}{0.727}  \\
\hline
\end{tabular}
\end{table}

\textbf{Tangent-space MSE vs.\ geodesic reconstruction loss.} Replacing the geodesic reconstruction loss with a tangent-space MSE reconstruction loss reduced performance on both datasets, from 0.95 to 0.91 top-1 accuracy on NTU-60 and from 0.740 to 0.718 in stroke $R^2$. This indicates that geodesic reconstruction provides a consistent, though modest, advantage over a Euclidean tangent-space reconstruction objective in the ES-VAE setting.

\section{Broader Impact}\label{sec:impact}
The experiments and modeling of elastic shape latent space in this work offer new ways for skeleton pose trajectory processing and contributes to the advancement of activity recognition and motion assessment of patients and different participants. It may have positive societal impact by improving interpretable analysis of skeleton pose trajectories for rehabilitation, activity recognition, and automated motion assessment. In clinical settings, compact latent representations of gait may help clinicians quantify mobility impairment, monitor rehabilitation progress, and identify motion patterns that are difficult to summarize manually. However, if used without consent, skeleton motion data could reveal sensitive health or activity information about a person. Therefore, this type of model should be used carefully, with privacy protection, clinical oversight, and validation on diverse groups of participants.

\end{document}